\documentclass{egpubl}
\usepackage{pg2026}

\SpecialIssuePaper
\CGFStandardLicense

\usepackage[T1]{fontenc}
\usepackage{dfadobe}
\usepackage{microtype}
\usepackage{graphicx}
\usepackage{subcaption}
\usepackage{booktabs}
\usepackage{amsmath}
\usepackage{amssymb}
\usepackage{mathtools}
\usepackage{multirow}
\usepackage[table]{xcolor}
\usepackage{algorithm}
\usepackage{algorithmic}
\usepackage{cite}
\usepackage{url}
\usepackage{egweblnk}
\BibtexOrBiblatex
\electronicVersion
\PrintedOrElectronic

\title[MG-Cache for Diffusion Transformer Inference]%
      {MG-Cache: Multi-Granularity Caching for Accelerating Diffusion Transformer Inference}

\author[Zesen Chen, Youxuan Xu, Shigang Li]%
      {\parbox{\textwidth}{\centering Zesen Chen\orcid{0009-0009-2133-4396}, Youxuan Xu\orcid{0009-0007-1698-3565}, Shigang Li\thanks{Corresponding author: shigangli.cs@gmail.com}\orcid{0000-0003-0022-7865}}\\%
       \parbox{\textwidth}{\centering Beijing University of Posts and Telecommunications, China}}

\begin{document}
\maketitle
\begin{abstract}
Diffusion Transformer (DiT) models, while having become a powerful backbone for generative synthesis, suffer from high inference cost on generating high-resolution images. Diffusion caching has emerged as a promising acceleration technique to reduce inference cost. However, existing caching strategies fail to accurately identify the varying importance of different computational modules and rarely coordinate cache reusing across the timestep and layer granularities, making it difficult to control error accumulation. To address these limitations, we propose multi-granularity caching (MG-Cache) for DiTs, a novel and fully training-free approach to exploit computational redundancy both on timesteps and Transformer layers. Our core innovation contains a two-stage caching strategy: an offline timestep profiling analysis that identifies redundant timesteps for cache reuse, followed by an adaptive layer-wise caching strategy that combines input stability with importance of each Transformer layer to selectively cache stable Attention and MLP submodules. Extensive experiments on mainstream open-source DiT models validate the effectiveness of our approach, achieving significant acceleration while maintaining high fidelity to the uncached output, with up to a 4.18$\times$ acceleration on HunyuanVideo.

\end{abstract}

% \begin{keywords}
% Diffusion models, Transformers, inference acceleration, caching, image and video generation
% \end{keywords}

\section{Introduction}
\begin{figure}[h]
    \centering
    \includegraphics[width=\linewidth]{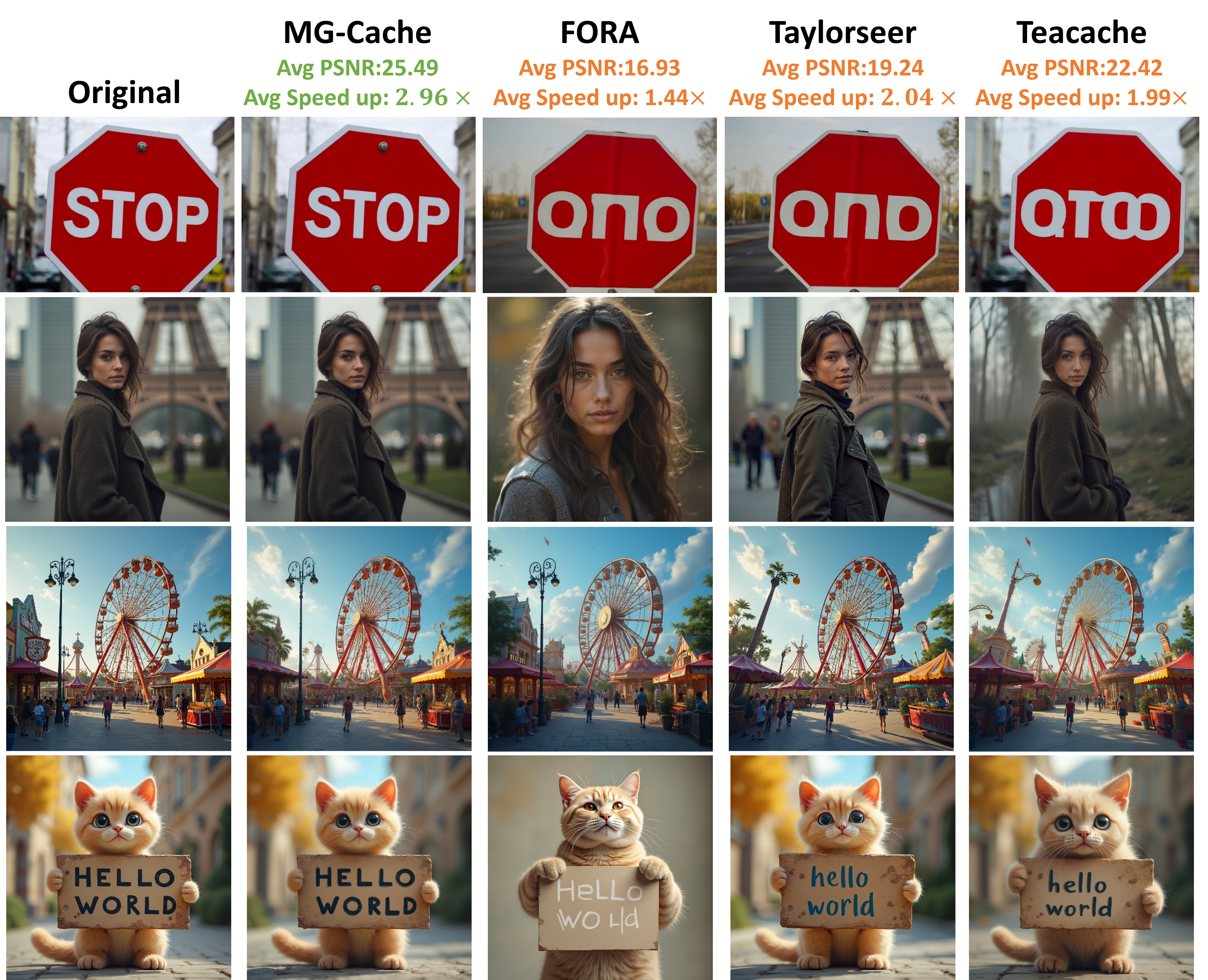}
    \caption{\textbf{Visual quality comparison.} While achieving a higher speedup than baselines, our method also maintains superior fidelity to the original output. For instance, in the first row (Prompt 1), other methods exhibit blurring on the text "stop," whereas our method preserves text sharpness.}
    \label{fig:image_accuracy}
\end{figure}
In recent years, multimodal artificial intelligence has progressed rapidly, with diffusion models \cite{song2020denoising, ho2022video} achieving breakthroughs in image synthesis and video generation tasks. Diffusion models now surpass generative adversarial networks (GANs) in both perceptual quality and generative fidelity \cite{dhariwal2021diffusion}. With the emergence of Diffusion Transformers (DiT) \cite{dosovitskiy2020image, peebles2023scalable}, diffusion frameworks have been tightly integrated with Transformer architectures, getting more expressive cross-modal representations and eliminating the structural limitations of U-Net-based encoder-decoder designs \cite{yang2024cogvideox, cao2025hunyuanimage}. 

Despite these advantages, the introduction of full 3D self-attention dramatically increases inference cost. Since attention computation grows quadratically with sequence length, high-resolution image generation and long-duration video synthesis all lead to extremely long token sequences in the latent space \cite{labs2025flux, wan2025wan, kong2024hunyuanvideo}, making attention the primary computational bottleneck in DiT inference. As generative models are increasingly deployed in industrial content creation and intelligent manufacturing \cite{li2024distrifusion, fang2024pipefusion}, the high computational overhead of diffusion inference has become a critical barrier to broader adoption.

To alleviate this issue, recent research has explored several acceleration strategies that improve inference efficiency. Among these approaches, diffusion caching has emerged as a promising direction \cite{wimbauer2024blockcache, kahatapitiya2025adaptive}. These training-free methods are grounded in a common observation that during iterative denoising, denoising steps that are adjacent or close to each other often exhibit highly similar internal feature representations, implying substantial computational redundancy. By caching intermediate features at earlier steps and reusing them at later steps, diffusion caching can reduce computation and accelerate generation efficiently. 

Although these methods show effectiveness on certain models, there are still notable limitations. Existing caching approaches fail to accurately identify the varying importance of individual timesteps and blocks. And most of them typically cache and reuse the model output solely at the timestep level. By treating the entire transformer network as a black box, these methods cannot exploit the inherent computational redundancy within the model's internal structure. Moreover, some methods adopt solely layer-granularity caching \cite{wimbauer2024blockcache, liu2025smoothcache}, but fail to effectively control error accumulation. Recent hybrid methods such as MixCache\cite{wei2025mixcache} combine multiple granularities, yet they still choose a single granularity per timestep via similarity- and perturbation-based criteria without a sensitivity-aware error budget, so the complementary redundancy across timesteps and layers is exploited separately. Consequently, pursuing higher inference speedup leads to a sharp degradation in image quality, and the achievable speedup of such approaches remains bounded.

\begin{figure*}[t]
    \centering
    \includegraphics[width=\linewidth]{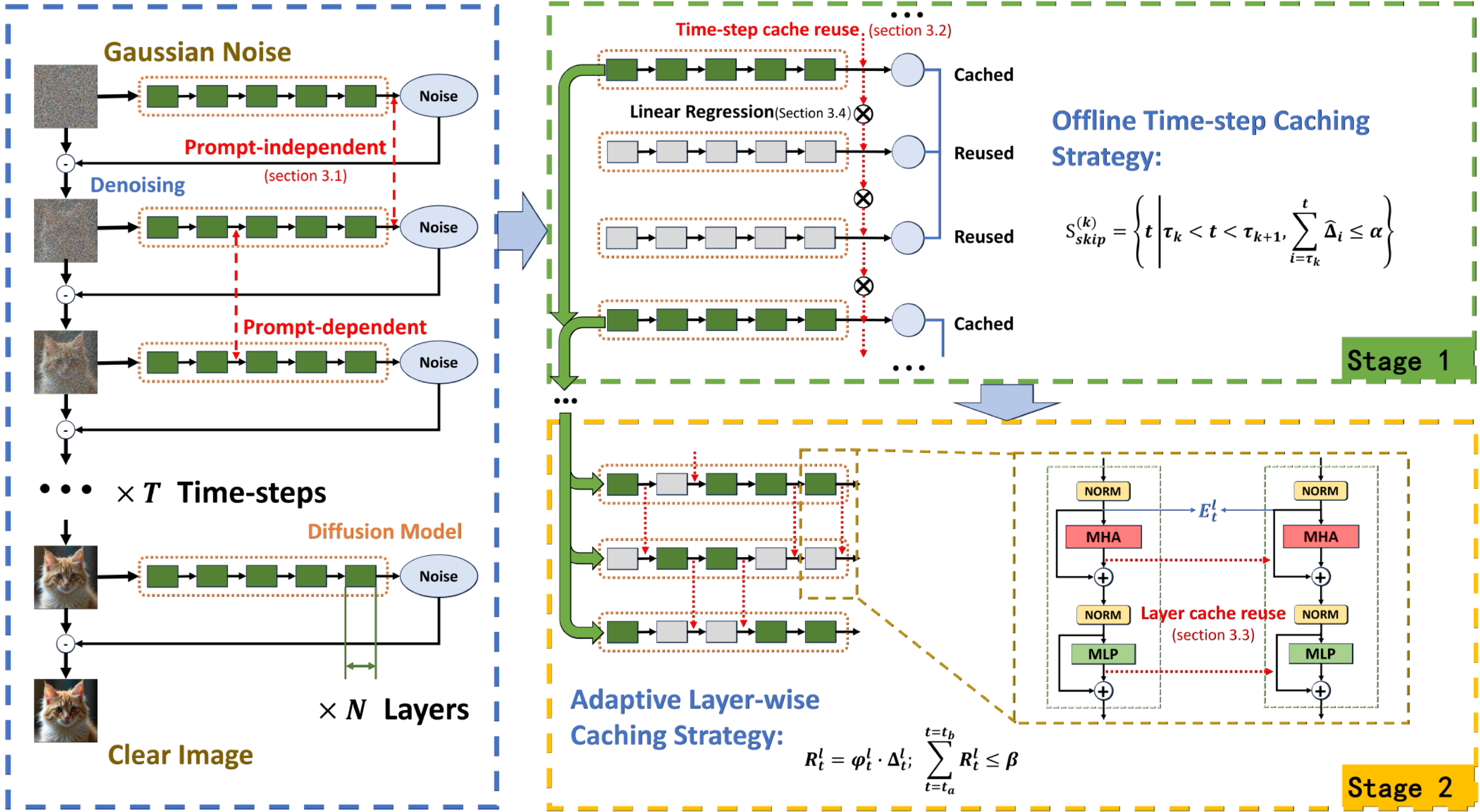}
    \caption{\textbf{Overview of the Multi-Granularity Cache (MG-Cache).} We design a two-stage strategy to exploit computational redundancy during inference. \textbf{Stage 1} performs a static offline analysis to identify and preserve critical timesteps based on a timestep feature table $[\hat{M}_t]$ constructed from a small batch of calibration dataset. \textbf{Stage 2} introduces a dynamic layer-wise caching mechanism that computes layer importance scores and input stability to adaptively determine whether to reuse cached features.}
    \label{fig:overview}
\end{figure*}

To address these limitations, this study proposes MG-Cache, a more comprehensive and multi-granularity caching method that synergistically combines timestep caching and layer-wise caching mechanisms. Specifically, based on our observations, the evolution patterns of timestep cache remain largely consistent across different generation processes, whereas layer-wise caches exhibit strong prompt-dependent behaviors. Motivated by this, we propose a two-stage strategy that integrates offline timestep analysis with online adaptive layer-wise caching as shown in Figure \ref{fig:overview}. In the first stage, we perform an offline analysis to identify a subset of timesteps in which the full computations can be safely skipped by reusing cached features. In the second stage, we introduce an online criterion based on the stability of the layer's input and importance scores. During inference, it adaptively determines which attention or MLP submodules can be cached at the remaining timesteps. Furthermore, we integrate a linear regression mechanism into the timestep cache to mitigate the errors introduced by straightforward cache reuse. This multi-granularity caching mechanism enables further redundancy exploitation while minimizing perceptual drift in visual quality. In addition, we conducted extensive experiments and ablation studies to demonstrate the effectiveness of our method.

The main contributions of our work are summarized as follows:
\begin{itemize}
    \item We propose MG-Cache, a novel multi-granularity caching framework for Diffusion Transformers. Unlike existing caching methods that reuse features at a single granularity or select one granularity per step, MG-Cache jointly exploits the complementary redundancy across both timesteps and Transformer layers within the same generation, enabling more effective acceleration while maintaining high image fidelity.
    \item We introduce a two-stage caching strategy that integrates offline timestep profiling with online adaptive layer-wise caching. The offline stage identifies redundant timesteps whose computations can be safely skipped, while the online stage dynamically selects stable Attention and MLP submodules based on input stability and importance scores, effectively controlling error accumulation during inference.
    \item We decompose the risk of dynamic layer-cache reuse into two complementary factors: a low-prompt-dependence robustness term of the downstream subnetwork, estimated offline through layer importance scores, and a prompt-dependent activation-drift term, measured online during inference. This decomposition explains why offline sensitivity profiling and online adaptive layer caching are logically consistent and mutually reinforcing.
    \item We further enhance timestep caching with a lightweight linear regression compensation mechanism. This mechanism mitigates the distribution shift introduced by cache reuse. Extensive experiments and ablation studies on HunyuanVideo and FLUX demonstrate that MG-Cache achieves up to a 4.18$\times$ speedup while maintaining high fidelity to the uncached output, with higher speedups than existing diffusion caching methods.
\end{itemize}

\section{Related Work}

\subsection{Diffusion Transformer}
 Diffusion Models have demonstrated unprecedented capabilities in high quality image synthesis. Fundamentally, these models operate by iteratively corrupting data with Gaussian noise and training a neural network to reconstruct the original signal \cite{ho2020denoising} through a reverse denoising process:
 \begin{equation}
     q(X_t|X_{t-1})=N(X_t;\sqrt{1-\beta_t}X_{t-1},\beta_{t}I)
 \end{equation}
The U-Net architecture has served as the standard for this denoising task. But the Transformer architecture has recently challenged this convention. For a mainstream Diffusion Transformer model, the noisy image $X_t$ at each step is first projected into the latent space via VAE and embedded into 1D token sequence $x_t^0$. This sequence then passes through n transformer layers and final linear layer to produce the predicted noise embedding for timestep t \cite{wan2025wan, kong2024hunyuanvideo, labs2025flux}. This can be expressed as:
\begin{equation}
    x_t^{l+1}=MLP(MHA(x_t^l))
\end{equation}
\begin{equation}
    M_t=\text{final linear}(x_t^n)
\end{equation}
 
Meanwhile, the Transformer architecture consists of more regular and homogeneous layers, making layer-wise caching in DiT models more straightforward than in U-Net-based architectures.

\subsection[Cache-based Acceleration for Diffusion Models]{Cache-based Acceleration for Diffusion Models}
Diffusion caching accelerates inference by storing intermediate features at selected steps and reusing them at later steps, a principle first exploited in U-Net-based diffusion models \cite{ma2024deepcache}. Along the timestep dimension, FORA \cite{selvaraju2024fora} recomputes at fixed intervals and reuses cached features in between, TeaCache \cite{liu2025teacache} skips steps based on the similarity of timestep-modulated first-layer features, and TaylorSeer \cite{liu2025taylorseer} fits the next-step output from a fixed number of past caches. Along the network dimension, PAB \cite{zhao2024real} adapts to the varying temporal stability of attention computations, SmoothCache \cite{liu2025smoothcache} applies a static uniform layer-cache schedule, and learned and token-level policies have also been explored \cite{ma2024learning,zou2024accelerating}. These methods restrict reuse to a single granularity and do not coordinate coarse timestep reuse with fine-grained layer reuse within one generation.

Recent hybrid and calibration-based methods improve the reuse operation itself. MixCache \cite{wei2025mixcache} integrates step, CFG, and block-level caching, yet it selects one granularity per timestep online using similarity- and perturbation-based heuristics without a sensitivity-aware error budget, so error accumulates under high speedup and complementary redundancy across timesteps and layers remains unexploited. Increment-calibrated caching \cite{chen2025incrementcalibrated} refines the reused values with low-rank calibration derived from the pre-trained model, without changing when reuse is allowed.

\subsection{Acceleration for Diffusion Models}

Existing research on accelerating diffusion models generally falls into several major directions. Some methods utilize sparse attention, such as Sparse VideoGen \cite{xi2025sparse} and MInference \cite{jiang2024minference}, attempt to reduce computational complexity by utilizing local or dynamic sparsity patterns derived from attention heat maps. However, these approaches often rely on specific hardware platforms or customized CUDA kernels to realize actual speed gains, which limits their portability and general applicability. Alternatively, Distillation techniques or other post-training paradigms have been adapted to compress the iterative denoising process into fewer steps \cite{salimans2022progressive, meng2023distillation}. Despite their potential for strong acceleration, they typically incur substantial post-training costs and are heavily dependent on the scale and quality of training data. In contrast, MG-Cache is training-free and accelerates the original model directly by reusing its intermediate features, without modifying the generation process. Similarly, quantization methods \cite{lin2024awq, frantar2022gptq} aim to compress model parameters and activations to low bit-widths, requiring additional Quantization-Aware Training to prevent accuracy loss, and fail to effectively accelerate image inference, as the overhead of handling large activation tensors can offset the benefits of parameter compression. Consequently, critical limitations still persist within these acceleration paradigms in the acceleration of diffusion transformers.

\section{Method}
We propose a novel two-stage caching strategy tailored for Diffusion Transformers (DiTs). Specifically, it contains two distinct stages, offline-analysis based timestep caching strategy in stage 1, and online adaptive layer-wise caching in stage 2. Furthermore, to mitigate the offset errors introduced by straightforward cache reuse, we integrate a linear regression fitting mechanism into timestep cache.

\subsection{Prompt-dependence of Timestep and Layer Cache}
\begin{figure}[tb]
    \centering
    \includegraphics[width=\linewidth]{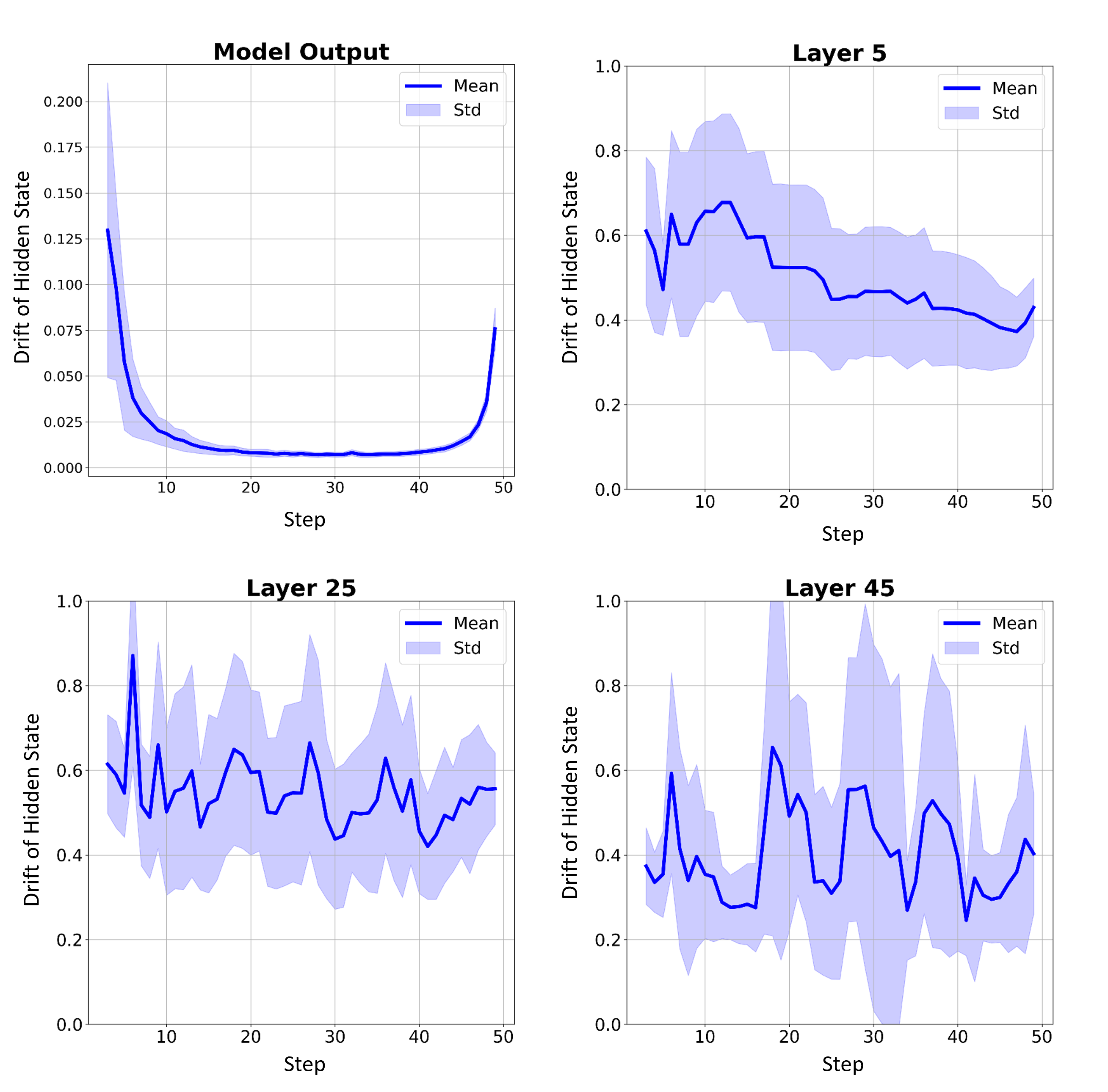}
    \caption{\textbf{Different evolution patterns of intermediate features at each timestep for different prompts.} The final model output predominantly exhibits \textbf{prompt-independent} patterns, whereas the intermediate features within layers display \textbf{prompt-dependent} patterns.}
    \label{fig:prompt_dependent}
\end{figure}
During repeated denoising inference in DiT, not only the final model output but also the intermediate features, such as attention and MLP outputs, change smoothly as the denoising steps progress. Based on the HunyuanVideo T2V model, we analyzed the evolution patterns of intermediate features at each timestep for different prompts by utilizing a small batch of calibration data. Crucially, our analysis reveals a significant divergence between the prompt dependence of timestep cache and that of layer cache. 

To analyze prompt dependence, we first quantify the magnitude of the raw drift of intermediate features across denoising steps. Let $F_t^l$ denote either the final model output at timestep $t$ or an intermediate feature in layer $l$. We define the normalized hidden-state drift between adjacent timesteps by Eq.~\ref{eq:drift}. Caching the final model output enables skipping of the entire timestep (\textbf{Timestep Cache}), while caching intermediate computation results, such as the outputs of Attention and MLP, allows for fine-grained layer-wise acceleration (\textbf{Layer Cache}).
\begin{equation}
    \Delta_t^l = \frac{\|F_t^l-F_{t-1}^l\|_1}{\|F_t^l\|_1 + \varepsilon}
    \label{eq:drift}
\end{equation}
\begin{equation}
    cv=\frac{\mathrm{std}(d_t^l)}{\mathrm{mean}(d_t^l)}
    \label{eq:cv}
\end{equation}
As illustrated in Figure \ref{fig:prompt_dependent}, where the solid line represents the mean value and the shaded area denotes the standard deviation on different prompts, the prompt dependence of model output and intermediate features is highly asymmetric. The standard deviation for the final model output is remarkably low, and the evolution curves remain tightly clustered across prompts, suggesting a low prompt-dependence. In contrast, the intermediate layer features show substantially larger cross-prompt variation, indicating a high prompt-dependence.

Furthermore, we utilized the Coefficient of Variation (CV) to quantify this prompt dependency by Eq.~\ref{eq:cv}. As shown in Table~\ref{tab:prompt_dependent}, the results confirm this difference in prompt dependency on timestep cache and layer cache. It suggests that timestep caching patterns can be determined via offline analysis, whereas layer-wise caching requires an online dynamic strategy to adaptively cache layers based on real-time stability evaluation. Motivated by these observations, we propose a two-stage strategy: offline-analysis based timestep caching in Stage 1, and online adaptive layer caching in Stage 2. The details of this implementation are elaborated in the following sections.

\begin{table}[h]
    \centering
    \caption{Prompt dependence of timestep and layer features.}
    \begin{tabular}{l|p{1.2cm} p{1.2cm} p{1.2cm} p{1.2cm}}
        \hline
    & \textbf{Model Output} & \textbf{Layer-15} & \textbf{Layer-35} & \textbf{Layer-55}  \\
        \hline
        \textbf{Std}     & \textbf{0.0114}       & 0.1952   & 0.1903   & 0.1582    \\
        \textbf{CV}      & \textbf{0.1504}       & 0.3419   & 0.3345   & 0.4195   \\
        \hline
    \end{tabular}
    \label{tab:prompt_dependent}
\end{table}
\subsection{Offline Timestep Caching Strategy}
Based on the prompt-dependence analysis above, the first stage of MG-Cache constructs a prompt-shared timestep schedule by accumulating a segment-wise error budget along the denoising trajectory. The timestep skipping process is viewed as allocating a finite reuse budget along the denoising trajectory. A timestep is skipped only when the accumulated risk since the most recent preserved step remains below a threshold $\alpha$; once the budget is exhausted, the current step is recomputed and becomes the new anchor.

In practice, we estimate this risk from the drift of the final model output between adjacent timesteps and aggregate it along the trajectory. Let $\tau_k$ denote the most recent preserved timestep before a candidate skipped segment. The hyperparameter $\alpha$ controls the cache rate along the timestep dimension. Therefore, the offline stage constructs a prompt-shared timestep policy in which each skip segment is selected under an explicit error budget, which serves as the coarse-grained backbone of MG-Cache.

\begin{equation}
    S_{\mathrm{skip}}^{(k)}=\left\{t \mid \tau_k < t < \tau_{k+1},\ \sum_{i=\tau_k}^{t}\hat{\Delta}_i \le \alpha\right\}
    \label{eq:stage1}
\end{equation}
The schedule $\hat{\Delta}_t$ is calibrated on a small batch of prompts, and the offline analysis strategy then identifies skipped timestep segments according to Eq.~\ref{eq:stage1}. During inference, if the current timestep belongs to the skipped timestep set, the timestep cache is reused to skip the entire denoising computation; otherwise, the process proceeds with the adaptive layer-wise caching strategy.

\subsection{Adaptive Layer-wise Caching Strategy}
As demonstrated in our previous analysis, it is necessary to adopt a dynamic and adaptive layer-wise caching mechanism. In the second stage, we maintain cached intermediate features for the remaining timesteps and estimate the risk of real-time reuse at each layer. A straightforward policy would reuse the cache whenever the current layer input is sufficiently close to the cached one, which can be expressed by Eq.~\ref{eq:naive}. However, empirical experiments indicated that this naive strategy produces unsatisfactory performance. This criterion ignores the fact that equal-magnitude perturbations can propagate very differently through different layers.
\begin{equation}
    \|x_t^l-x_{t-1}^l\|_1<\epsilon
    \label{eq:naive}
\end{equation}
\begin{algorithm}[tb]
    \caption{Layer Importance Scores Calculation}
    \label{alg:stage2}
    \begin{algorithmic}
        \STATE \textbf{Input:} DiT blocks $\{Block^l\}_{l=1}^{n}$, total inference steps $T$, fixed perturbation $\varepsilon$ with $\mathrm{mean}(|\varepsilon|)=0.1$
        \FOR{$t=0$ to $T$}
            \STATE Compute original model output $M_t$
            \FOR{$l=1$ to $n$}
                \STATE Compute intermediate feature $F_t^l$ (output of MHA or MLP) for block $l$
                \STATE Inject perturbation into the current feature: $\tilde{F}_t^l = F_t^l + \varepsilon$
                \STATE Perform forward propagation from layer $l$ to obtain perturbed output $\hat{M}_t^{\,l}$
                \STATE $\varphi_t^l=\|M_t-\hat{M}_t^{\,l}\|_1$
            \ENDFOR
        \ENDFOR
        \STATE \textbf{Output:} layer importance scores $\{\varphi_t^l\}$ for each layer $l$ and timestep $t$
    \end{algorithmic}
\end{algorithm}

We consider the sensitivity of each layer and the magnitude of activation drift separately. Assuming $0-T$ indicates the index of all remaining time steps at this point, let $G_t^l$ denote the downstream subnetwork that maps the input of layer $l$ at timestep $t$, namely $x_t^l$, to the final model output $M_t$. Reusing cached features from timestep $t-1$ introduces a perturbation $\delta_t^l = x_t^l - x_{t-1}^l$. Under a first-order local linearization of $G_t^l$, the resulting output error satisfies:
\begin{equation}
    \|M_t(x_t^l) - M_t(x_{t-1}^l)\|_1 \leq \|J_t^l\|_{1}\cdot\|\delta_t^l\|_1+O(\|\delta_t^l\|_1^2)
    \label{eq:layerSensitivity}
\end{equation}
where $J_t^l=\partial G_t^l/\partial x_t^l$ denotes the Jacobian of $G_t^l$ with respect to $x_t^l$. This decomposition separates the reuse error into a layer-sensitivity term $\|J_t^l\|_{1}$ and a drift term $\|\delta_t^l\|_1$. The sensitivity term is primarily determined by the layer position and denoising stage and is therefore much more stable across prompts, which is also supported by our additional statistics in Section~4.8. Meanwhile, the drift term is prompt-dependent and must be measured online. Consequently, offline importance estimation and online adaptive caching are not contradictory, as they estimate different factors of the same error bound.
\begin{equation}
    R_t^l = \varphi_t^l \cdot \Delta_t^l
    \label{eq:layerError}
\end{equation}
\begin{equation}
    \sum_{i=t_a^l}^{t}R_i^l\leq\beta
    \label{eq:layerBudget}
\end{equation}
MG-Cache scores each potential reuse operation by a sensitivity-aware risk proxy, following Eq.~\ref{eq:layerError}, where $\Delta_t^l$ is the online normalized activation drift and $\varphi_t^l$ is the layer importance score. The quantity $\varphi_t^l$ estimates how strongly a perturbation at layer $l$ and timestep $t$ affects the final denoising output; in the first-order view, it serves as a practical surrogate for the downstream Jacobian magnitude $\|J_t^l\|_{1\rightarrow 1}$. We obtain it by injecting a fixed perturbation into the intermediate feature and measuring the induced deviation in the final model prediction, as shown in Algorithm~\ref{alg:stage2}. 

In this way, $\varphi_t^l$ captures the intrinsic robustness of a certain layer, while $\Delta_t^l$ captures the prompt-dependent hidden-state evolution observed during real-time inference. Similarly to Stage 1, the accumulated risk in Eq.~\ref{eq:layerBudget} is introduced to prevent excessive amplification, where $t_a^l$ denotes the most recent recomputed step of layer $l$ (the anchor step, where the accumulator resets). The hyperparameter $\beta$ controls the cache rate within the remaining timesteps.

\begin{table}[h]
    \centering
    \caption{Comparison with adaptive layer-wise caching and naive strategy(HunyuanVideo $544\times960$ 129frames)}
    \begin{tabular}{l|p{0.8cm} p{0.8cm} p{1cm}|p{1.2cm}}
        \textbf{Methods} &\textbf{PSNR}&\textbf{SSIM}&\textbf{LPIPS}$\downarrow$&\textbf{FLOPs(p)}  \\
        \hline
         Naive Strategy  & 17.07 & 0.61 & 0.43 & 70.21         \\
        With Scores $\varphi_t^l$ & \textbf{27.44} & \textbf{0.86} & \textbf{0.13} & \textbf{67.74}  \\
        \hline
    \end{tabular}
    \label{tab:layer_important_ablation}
\end{table}

In practical implementation, the importance scores $[\varphi_t^l]$ are computed offline on a calibration dataset. We observed that the values of $[\varphi_t^l]$ remain highly consistent across different prompts, indicating that the layer importance score is largely prompt-independent. This further confirms that the importance scores capture the intrinsic, prompt-invariant robustness of each layer, while the stability of layer inputs is prompt-dependent and thus shows adaptivity during inference.

We conducted ablation studies to compare the naive strategy against our adaptive importance-based approach. As shown in Table \ref{tab:layer_important_ablation}, the importance-based adaptive scoring strategy achieves a significant improvement in image fidelity, while both strategies attain a similar reduction in computational load (FLOPs).

\subsection{Linear Regression on Timestep Cache}
\begin{figure}[htbp]
    \centering
    \includegraphics[width=\linewidth]{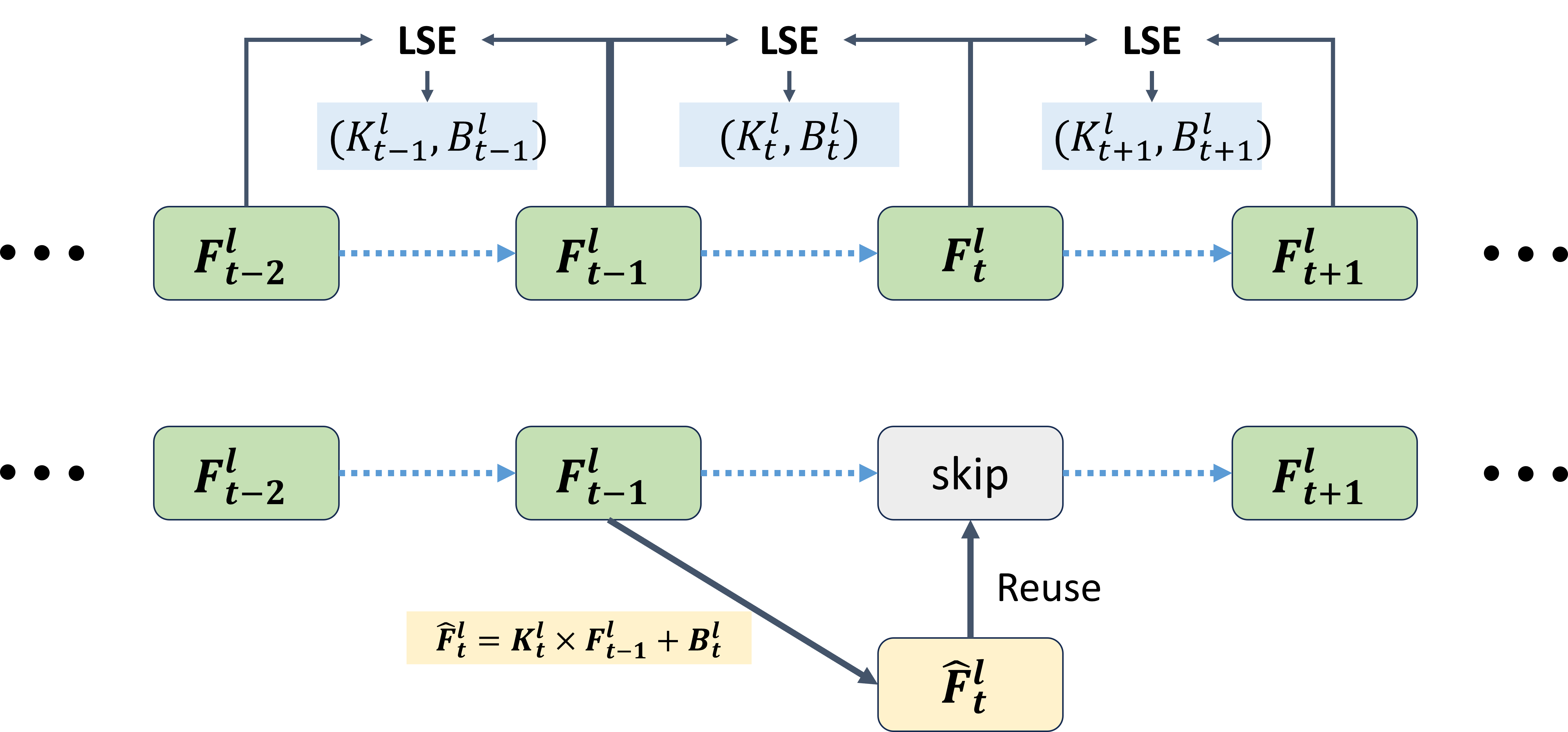}
    \caption{\textbf{Linear Regression.} Instead of directly caching and reusing, a linear regression transformation is applied for cached features.}
    \label{fig:linear}
\end{figure}
To further mitigate the feature drift error induced by direct cache reuse, we introduce a simple linear regression mechanism on timestep caching. Specifically, for the intermediate features at a given timestep $t$ and channel $c$, by using Least Squares Estimation (LSE), we utilize feature pairs $(F_{c,t}^l, F_{c,t-1}^l)$ to derive a linear mapping as follows:
\begin{equation}
    \hat{F}_{(c,t)}^l=K_{(c,t)}^l\cdot F_{(c,t-1)}^l+B_{(c,t)}^l
    \label{eq:linear}
\end{equation}
The coefficients $K_t^l$ and $B_t^l$ are first estimated on a small batch of calibration dataset. During inference, the fitted cache values are used to replace the directly reused cached results, reducing the bias introduced by simple cache reuse. This approach effectively recovers the performance of caching mechanism without fine tuning.
\begin{figure}[htbp]
    \centering
    \begin{subfigure}{0.495\linewidth}
        \centering
        \includegraphics[width=\linewidth]{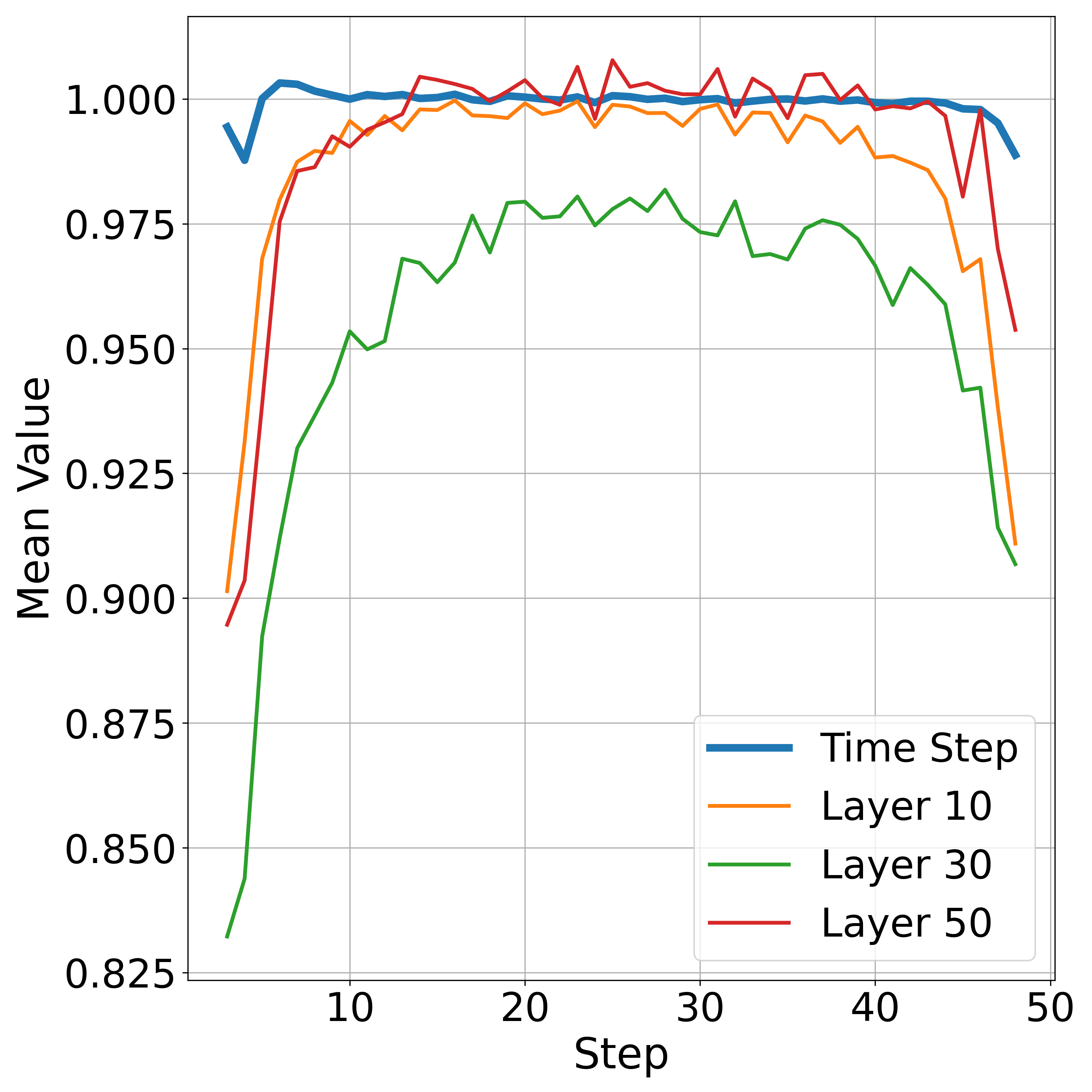}
        \caption{Mean Value of K}
    \end{subfigure}
    \hfill
    \begin{subfigure}{0.495\linewidth}
        \centering
        \includegraphics[width=\linewidth]{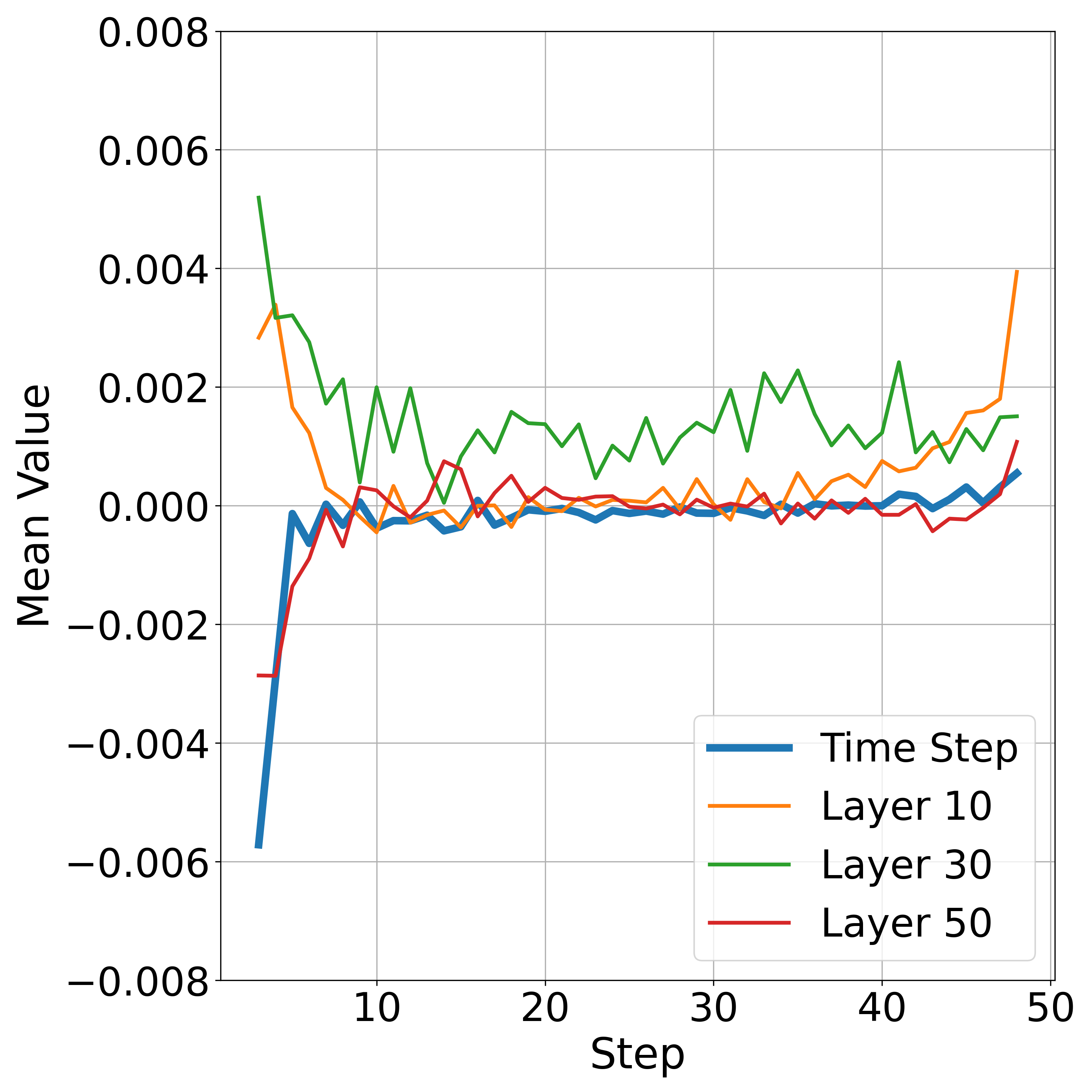}
        \caption{Mean Value of B}
    \end{subfigure}
    \caption{\textbf{Mean value of linear regression coefficients.} Different value of coefficients in each layer cache and timestep cache on HunyuanVideo. }
    \label{fig:mean_value_KB}
\end{figure}

We validated the efficacy of this design through ablation studies as shown in Table \ref{tab:ablation-study}, which demonstrate that this lightweight linear regression mechanism significantly restores image generation fidelity. Furthermore, we analyzed the distribution of the regression coefficients $K_c$ and $B_c$ of different layers. As shown in Figure \ref{fig:mean_value_KB}, as for timestep cache, $K$ are consistently concentrated around 1, and $B$ remains close to 0, indicating that the linear regression mainly performs as precise lightweight compensation. In contrast, as for layer cache, the coefficients $K$ deviate more significantly from 1, implying a more aggressive transformation. Empirically, we found that applying linear regression compensation to layer cache provides limited or even negative benefits. Therefore, in practice, we only employ the linear regression compensation for timestep cache.

The complete inference procedure of MG-Cache for generating a single video is summarized in Algorithm~\ref{alg:inference}. 
\begin{algorithm}[tb]
    \caption{Inference procedure of MG-Cache for a single video.}
    \label{alg:inference}
    \begin{algorithmic}
        \STATE \textbf{Input:} prompt $p$; initial latents $x_0$; offline skip schedule $\mathcal{S}$ (Eq.~\ref{eq:stage1}); layer importance scores $\varphi_t^l$; regression coefficients $K_t, B_t$; layer budget $\beta$.
        \STATE \textbf{Initialize:} timestep cache $M_{\mathrm{cache}}$; per-layer references $\bar{x}^l$ and cached outputs $h_{\mathrm{cache}}^l$; risk accumulators $E^l\leftarrow 0$ (caches populated at the first computed step).
        \FOR{$t=0$ to $T$}
            \IF{$t\in\mathcal{S}$}
                \STATE $M_t\leftarrow K_t\cdot M_{\mathrm{cache}}+B_t$
            \ELSE
                \STATE Compute the conditioning embeddings and block inputs
                \FOR{each layer $l$ in the DiT}
                    \STATE $\Delta_t^l\leftarrow\mathrm{relL1}(\bar{x}^l, x_t^l)$; \quad update the reference $\bar{x}^l\leftarrow x_t^l$
                    \STATE $E^l\leftarrow E^l+\varphi_t^l\cdot\Delta_t^l$
                    \IF{$E^l<\beta$}
                        \STATE $h_t^l\leftarrow h_{\mathrm{cache}}^l$
                    \ELSE
                        \STATE Compute the layer outputs; refresh $h_{\mathrm{cache}}^l$; set $E^l\leftarrow 0$
                    \ENDIF
                \ENDFOR
                \STATE $M_t\leftarrow$ output of the final layer; \quad $M_{\mathrm{cache}}\leftarrow M_t$
            \ENDIF
            \STATE $x_{t+1}\leftarrow\mathrm{scheduler\_step}(x_t, t, M_t)$
        \ENDFOR
        \STATE \textbf{Return:} output video $\leftarrow\mathrm{decode}(x_T)$
    \end{algorithmic}
\end{algorithm}

\section{Experiment}
\begin{table*}[htb]
    \centering
    \caption{Quantitative results of generation quality and inference acceleration, evaluated on the \textbf{HunyuanVideo} and \textbf{FLUX.1-dev}}
    \begin{tabular}{l|c c c c|c c c}
        \hline
        \rowcolor{gray!20}
        \multicolumn{8}{c}{\textbf{HunyuanVideo-T2V}, $544\times960$p, 129 frames}\\
        \hline
         \multirow{2}{*}{\textbf{Methods}} & \multicolumn{4}{c|}{\textbf{Quality}} & \multicolumn{3}{c}{\textbf{Acceleration}}\\
         \cline{2-8}
         &\textbf{VBench Score}$\uparrow$&\textbf{PSNR}$\uparrow$&\textbf{SSIM}$\uparrow$&\textbf{LPIPS}$\downarrow$&\textbf{FLOPs(P)}$\downarrow$&\textbf{Latency(s)}$\downarrow$&\textbf{SpeedUp}$\uparrow$ \\
         \hline
    Original(50 steps)& 75.45\% & -     & -    & -   & 214.6299 & 982.03 & - \\
    \hline
         20 steps     & 75.21\% & 18.16 & 0.67 &0.38  & 85.8519 & 413.38 & 2.38$\times$ \\
         FORA($N=3$)   & 74.75\%  & 17.62 & 0.66 &0.40  & 101.6018 & 482.80 & 2.03$\times$ \\
   Teacache($\delta=0.4$)& 75.33\% & 23.99 & 0.79 &0.19  & 94.4372 & 457.01 & 2.15$\times$ \\
   MixCache(effi)& \textbf{75.43\%} & 25.95 & 0.83 & 0.16 & 111.23 & 503.1 & 1.98$\times$ \\
   MixCache($N=6$) & 73.94\% & 23.30 & 0.78 & 0.25 & 82.1 & 390.2 & 2.52$\times$ \\
   SmoothCache($\delta=0.16$) & 72.53\% & 23.10 & 0.76 & 0.34 & 122.18 & 586.6 & 1.67$\times$ \\
          \hline
    ours($\alpha=0.04$,$\beta=0.3$)&75.14\%&\textbf{27.44} &\textbf{0.86}&\textbf{0.13} & 67.7417 & 358.86 & 2.74$\times$ \\
    ours($\alpha=0.07$,$\beta=0.4$) &74.13\%& 25.77 & 0.84 &0.17  &\textbf{51.3572} &\textbf{283.28} & \textbf{3.47}$\times$\\
    \hline
    \end{tabular}\\[1\baselineskip]

    \begin{tabular}{l|c c c c|c c c}
        \hline
        \rowcolor{gray!20}
        \multicolumn{8}{c}{\textbf{HunyuanVideo-T2V}, $480\times640$p, 65 frames}\\
        \hline
         \multirow{2}{*}{\textbf{Methods}} & \multicolumn{4}{c|}{\textbf{Quality}} & \multicolumn{3}{c}{\textbf{Acceleration}}\\
         \cline{2-8}
         &\textbf{VBench Score}$\uparrow$&\textbf{PSNR}$\uparrow$&\textbf{SSIM}$\uparrow$&\textbf{LPIPS}$\downarrow$&\textbf{FLOPs(P)}$\downarrow$&\textbf{Latency(s)}$\downarrow$&\textbf{SpeedUp}$\uparrow$ \\
         \hline
    Original(50 steps)&74.22\% & -     & -    & -   & 29.5335 & 160.42 & - \\
    \hline
         20 steps     &74.10\% & 18.48 & 0.67 & 0.31 & 11.8134 & 74.63 & 2.15$\times$ \\
         FORA($N=3$)   & 74.13\% & 18.09 & 0.64 & 0.33 & 19.2080 & 111.1853 & 1.44$\times$ \\
   Teacache($\delta=0.15$)    & 73.86\% & 23.20 & 0.78 & 0.20 & 12.9947 & 80.53 & 1.99$\times$ \\
  Taylorseer($N=3$,$O=1$)& \textbf{74.61\%} & 17.86 & 0.65 & 0.33 & 11.2227 & 78.73 & 2.04$\times$ \\
   MixCache(effi)& 73.84\% & 24.56 & 0.82 & \textbf{0.14} & 16.82 & 89.4 & 1.81$\times$ \\
   MixCache($N=6$) & 72.67\% & 22.72 & 0.75 & 0.25 & 13.57 & 73.7 & 2.2$\times$ \\
   SmoothCache($\delta=0.16$) & 71.72\% & 20.88 & 0.70 & 0.38 & 17.11 & 109.5 & 1.47$\times$ \\
         \hline
    ours($\alpha=0.04$,$\beta=0.3$)  & 74.58\% & \textbf{25.28 }& \textbf{0.82} & 0.15 & 9.3974 & 69.66 & 2.30$\times$ \\
    ours($\alpha=0.07$,$\beta=0.4$)  & 73.64\% & 24.45 & 0.79 & 0.18 & \textbf{7.1251} & \textbf{54.18} & \textbf{2.96}$\times$\\
    \hline
    \end{tabular}\\[1\baselineskip]
    
    \begin{tabular}{l|c c c c|c c c}
         \hline
         \rowcolor{gray!20}
         \multicolumn{8}{c}{\textbf{FLUX.1-dev}, $1024\times1024$p, 1 frames}\\
         \hline
         \multirow{2}{*}{\textbf{Methods}} & \multicolumn{4}{c|}{\textbf{Quality}} & \multicolumn{3}{c}{\textbf{Acceleration}}\\
         \cline{2-8}
         &\textbf{CLIP Score}$\uparrow$&\textbf{PSNR}$\uparrow$&\textbf{SSIM}$\uparrow$&\textbf{LPIPS}$\downarrow$&\textbf{FLOPs(T)}$\downarrow$&\textbf{Latency(s)}$\downarrow$&\textbf{SpeedUp}$\uparrow$ \\
             \hline
   Original(28 steps)& 29.95 & -     & -    & -   & 1948.89 & 9.5270 & - \\
   \hline
         16 steps    & 29.80 & 17.57 & 0.71 &0.33 & 835.30 & 5.6816 & 1.68$\times$ \\
         FORA($N=3$)& 29.72& 15.08 & 0.63 &0.47 & 1174.78 & 6.9931  & 1.36$\times$ \\
   Teacache($\delta=0.4$)&\textbf{29.91}& 19.08 & 0.73 &0.30 & 904.87&4.6634 & 2.04$\times$ \\
        Taylorseer($N=3$,$O=1$)& 29.75 & 19.34 & 0.72 &0.29 & 695.95 & 4.4205     &2.16$\times$ \\
        \hline
         ours($\alpha=0.18$,$\beta=0.25$)  & 29.90 & \textbf{23.18} & \textbf{0.83} &\textbf{0.19} & 759.55 & 3.9818 & 2.39$\times$ \\
         ours($\alpha=0.27$,$\beta=0.30$)  & 29.86 & 22.12 & 0.78 &0.27 &\textbf{620.25} & \textbf{3.5259} & \textbf{2.70}$\times$\\
         \hline
    \end{tabular}
    \label{tab:main_result}
\end{table*}
\subsection{Settings}

\textbf{Base Models.} To validate the effectiveness of MG-Cache, we selected mainstream open-source image and video generation models for evaluation, including FLUX and HunyuanVideo. Specifically, FLUX.1-dev serves as the text-to-image generation model, while HunyuanVideo is employed as text-to-video model. The resolution for FLUX is set to $1024\times1024$, and for HunyuanVideo, the resolution is configured at $544\times960$ with a frame count of 129, and $480\times640$ with a frame count of 65. The number of inference steps is set to 50 for HunyuanVideo, and 28 steps for FLUX.1-dev, according to the default configuration.

\textbf{Baselines.} We adopted FORA \cite{selvaraju2024fora}, TeaCache \cite{liu2025teacache} and TaylorSeer \cite{liu2025taylorseer} as baseline methods for comparison. The hyperparameters of FORA and Teacache remain default settings, and TaylorSeer reduces $N$ from 6 to 3 to match similar speedup. It is important to note that TaylorSeer incurs significant GPU memory overhead, as it requires storing the intermediate Attention and MLP outputs of all layers across multiple past time steps. Therefore, it fails to complete the $544\times960$ video inference due to GPU out-of-memory (OOM). We additionally include MixCache\cite{wei2025mixcache} (its official efficient configuration and a variant with the cache interval increased from 3 to 6) and SmoothCache\cite{liu2025smoothcache} (threshold $\delta=0.16$) under the same setting.

\textbf{Datasets and Metrics.} For image generation, following the methods in TaylorSeer, we utilized 200 prompts from DrawBench \cite{saharia2022drawbench} for evaluation. We used the CLIP Score to assess the semantic alignment between the generated images and the text prompts. For video generation, 200 prompts were randomly selected from the VBench \cite{huang2024vbench} test suite to generate videos for evaluation. The aesthetic quality of the generated videos was assessed by using the VBench score. Additionally, consistent with the methods in TeaCache, we utilized PSNR, SSIM, and LPIPS to quantify the fidelity of the accelerated images and videos relative to the original outputs.

\textbf{Implementation Details.} During timestep calibration and layer-importance estimation, we randomly sampled 50 prompts from the Penguin Video Benchmark \cite{kong2024hunyuanvideo} as calibration dataset. We found that the calibrated timestep schedule and the layer-importance scores both exhibit low cross-prompt variance, supporting the use of a shared offline sensitivity profile. All experiments were conducted in a PyTorch environment on NVIDIA H100 80GB GPUs. Flash Attention was enabled by default for all evaluated methods.

\subsection{Main Results}
\textbf{Acceleration Efficiency.} Table \ref{tab:main_result} presents the primary quantitative results, incorporating metrics for visual quality (VBench, CLIP, PSNR, SSIM, and LPIPS) and computational efficiency (FLOPs and Latency). Based on our multi-granularity caching method, we configured two settings on HunyuanVideo: \textbf{Slow Configuration} with $\alpha=0.04$ and $\beta=0.3$, \textbf{Fast Configuration} with $\alpha=0.07$ and $\beta=0.4$. Without requiring any retraining or fine-tuning, MG-Cache achieved total computational cost reductions of 68.43\% and 76.07\% respectively within slow and fast configuration. Specifically, in the HunyuanVideo (5s, 960p) generation scenario, we achieved a speedup of 3.47$\times$, reducing the generation time to 283.28 seconds. Similarly, our method also achieves the best quality scores and the highest speedup on the text-to-image model FLUX.1-dev. Moreover, as video resolution and duration increase, the acceleration advantage of our method becomes more pronounced, as shown in Figure \ref{fig:speedup}.

\textbf{Image and Video Quality.} Regarding visual quality, our \textbf{Slow} configuration achieves the best fidelity at $544\times960$ while running faster than every compared baseline, and our \textbf{Fast} configuration attains the highest speedup (3.47$\times$) with fidelity comparable to the strongest baseline. Moreover, both VBench score and CLIP score remain at a high level without a significant decrease. Notably, PSNR quantifies the pixel-level disparity, and our method achieves a peak value of 27.44 dB. Figure \ref{fig:image_accuracy} shows more details of image quality comparison, demonstrating that our method maintains the highest visual fidelity to the original outputs.

\textbf{Quantitative Comparison.} FORA relies on fixed-step skipping and recomputation, ignoring the varying importance of different denoising steps. Taylorseer also relies on fixed intervals for recomputation; although reducing intervals $N$ from 6 to 3, it suffers a significant degradation in PSNR accuracy. TeaCache fails to exploit the redundancy within the model, limiting its ability to achieve higher acceleration ratios without compromising image quality. MixCache selects a single granularity per timestep without coordinating each level caches, which limits its achievable speedup and makes quality degrade sharply when the cache interval is enlarged. SmoothCache relies solely on uniform layer-level reuse and cannot exploit timestep-level redundancy, resulting in significant quality degradation even at a modest speedup. In contrast, MG-Cache adopts a two-stage strategy that precisely analyzes the importance of each timestep and further reduces internal computational redundancy through adaptive layer-wise caching, enabling higher inference acceleration while preserving high-fidelity generation quality.

\subsection{Ablation Study}
To validate the effectiveness of individual components within our method, we conducted a series of ablation studies. Based on HunyuanVideo, all experiments were conducted at the resolution of $544\times960\times129$.

\textbf{Impact of Offline Timestep Analysis.} We assessed the necessity of our offline timestep analysis by replacing it with a fixed step interval skipping strategy. As shown in Table \ref{tab:ablation-study}, setting the cache step length $N=3$ and keeping other configurations unchanged results in a significant decline in both VBench scores and PSNR. This demonstrates that importance varies significantly across timesteps; consequently, the selection of cached steps should be driven by their contribution to the final image generation rather than a fixed step interval.
\begin{table}[tb]
    \centering
    \caption{Ablation Study of each component on HunyuanVideo}
    \begin{tabular}{p{1.5cm}|p{1.4cm}|p{1cm}p{1cm}|p{1.2cm}}
    \hline
        \multicolumn{2}{l|}{\textbf{Method}}&\textbf{VBench}&\textbf{PSNR}&\textbf{SpeedUp}  \\
        \hline
        \rowcolor{gray!20}
        \multicolumn{2}{l|}{MG-Cache(slow)}&75.14\%&27.44&\textbf{2.74$\times$} \\
        \multicolumn{2}{l|}{w/o Offline Analysis}&74.79\%&24.15& 2.41$\times$ \\
        \multicolumn{2}{l|}{w/o Linear Regression}&74.51\% &26.82 &2.68$\times$\\
        \hline
        \multirow{2}{=}{w/o Time- step cache}&$\beta=0.3$&75.67\%&29.49&$1.23\times$ \\
                                            &$\beta=0.4$&74.86\%&26.83&1.42$\times$ \\
        \hline
        \multirow{2}{=}{w/o Layer cache}&$\alpha=0.04$&75.21\%&28.65&$2.52\times$ \\
                                        &$\alpha=0.05$&74.92\%&27.33&2.70$\times$ \\
    \hline
    \end{tabular}
    \label{tab:ablation-study}
\end{table}

\textbf{Effectiveness of Linear Regression.} In this experiment, we analyzed the impact of linear regression for recovery mechanism. The results demonstrate that applying a simple linear regression fit on timestep cache significantly recovers final image fidelity without the need for any re-training or fine-tuning. Moreover, without linear regression, adaptive layer-wise caching skips fewer layers, leading to a slight drop in speedup.

\textbf{Only Adaptive Layer-wise Cache.} We evaluated a configuration that relies solely on layer caching across all de-nosing steps. The results indicate that the speedup decreases significantly without timestep cache. Although increasing $\beta$ from 0.3 to 0.4 allows a larger ratio of skipped layers (from 25.79\% to 33.44\%), both the generation quality and the speedup become inferior to our original method. Moreover, we observed that skipping a large number of layers within a single timestep introduces cumulative errors that exceed those caused by direct skipping the whole timestep. Therefore, integrating timestep cache alleviates cumulative errors in diffusion caching.

\textbf{Only Timestep Cache.} Finally, we excluded layer-wise caching to rely only on static timestep caching. The results show the overall acceleration drops without layer cache. Although raising $\alpha$ from 0.04 to 0.05 allows more timesteps to be skipped to compensate for the lost speedup, it causes VBench and PSNR to fall below our original method. This shows that integrating fine-grained layer-wise cache is necessary to achieve higher acceleration with preserved image quality.

\subsection{Efficiency with Different Sequence Length.}
\begin{figure}[h]
    \centering
    \includegraphics[width=\linewidth]{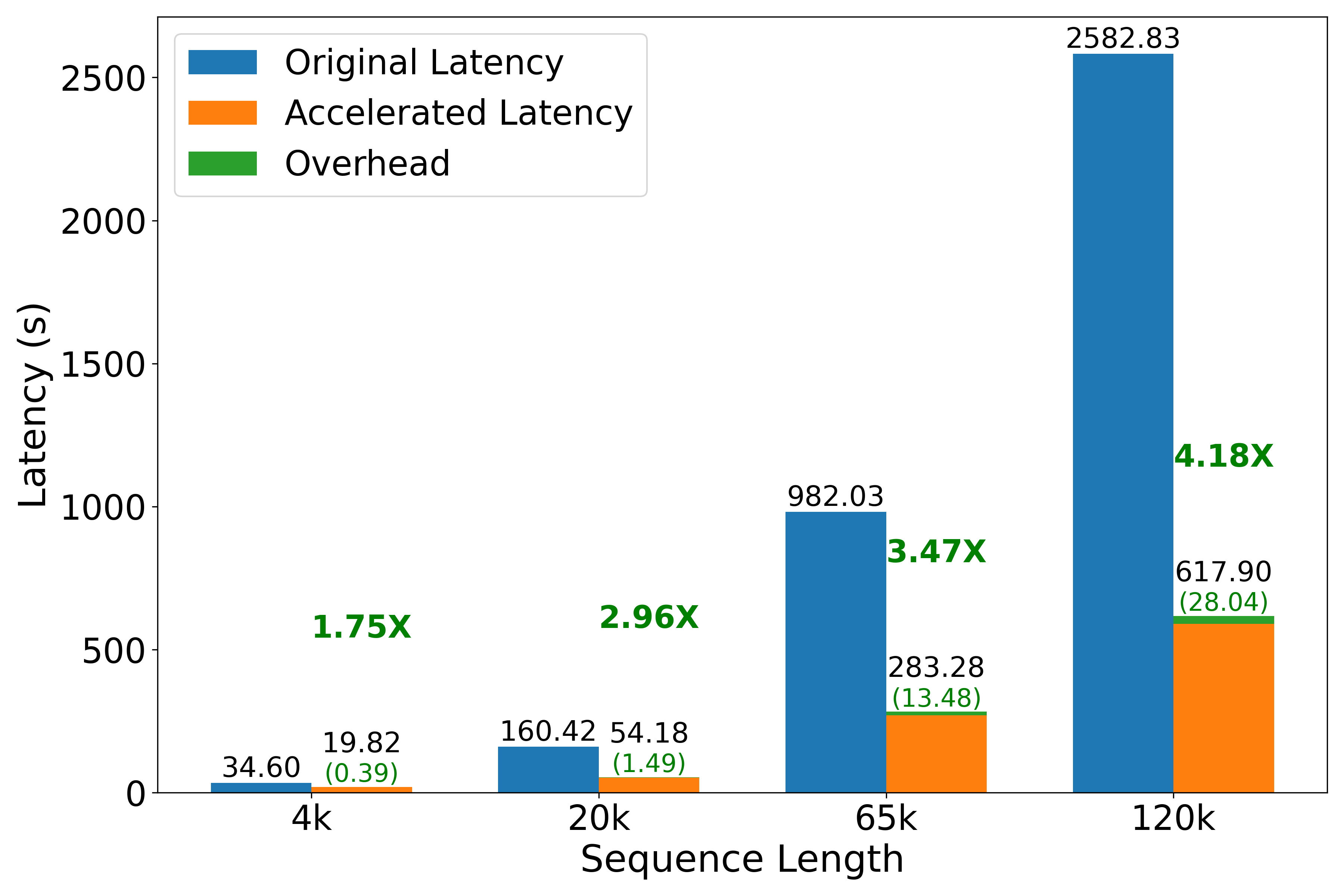}
    \caption{\textbf{Inference latency and added overhead with different resolution.} Since our method primarily targets compute-bound components such as self-attention and MLP modules through diffusion caching and selective skipping, the achievable speedup correspondingly improves with longer sequences, while the added overhead remains below 5\% of the total latency.}
    \label{fig:speedup}
\end{figure}
We evaluated the inference latency of DiT models across various sequence lengths on a single GPU. As illustrated in Figure \ref{fig:speedup}, our method achieves a speedup that increases with the token sequence length. Our multi-granularity caching method achieves a maximum speedup of $4.18\times$ on HunyuanVideo when generating 5-second videos at a resolution of $720 \times 1280$. 

Our approach maintains a fixed-size cache tailored to the sequence length. When the sequence length exceeds 40k tokens, the GPU memory (VRAM) becomes insufficient to store all the cache. To address this, we implemented a cache offloading mechanism for long sequences, moving the cache to system memory (RAM). During inference, we prefetch the cache required for the subsequent layer, effectively masking the latency of data transfer from system memory to GPU memory (including main results shown in Table \ref{tab:main_result}). Experimental results show that the additional overhead introduced by our method, including the offloading operations, remains below 5\% of the total inference latency. To identify the overhead, we measured the latency with $\beta=0$, where no layer reuse is applied and all caching, prefetching, and offloading mechanisms remain active. Under this setting, the additional latency remains below 5\% across all evaluated sequence lengths, as shown in Fig.~\ref{fig:speedup}. 

\subsection[Correlation between the Risk Proxy and the True Model Output Error]{Correlation between the Risk Proxy and the True Model Output Error}

The layer-wise reuse criterion should rank reuse operations by the error they actually introduce to the final output, rather than by the raw input drift alone. We therefore measure how strongly the sensitivity-aware risk proxy $R_t^l=\varphi_t^l\cdot\Delta_t^l$ (Eq.~\ref{eq:layerError}) and the raw activation drift $\Delta_t^l$ correlate with the true output error $M=\|M_t(x_t^l)-M_t(x_{t-1}^l)\|_1$ (the left-hand side of Eq.~\ref{eq:layerSensitivity}), over 1200 reuse decisions collected from 20 prompts.

\begin{figure}[tb]
    \centering
    \includegraphics[width=\linewidth]{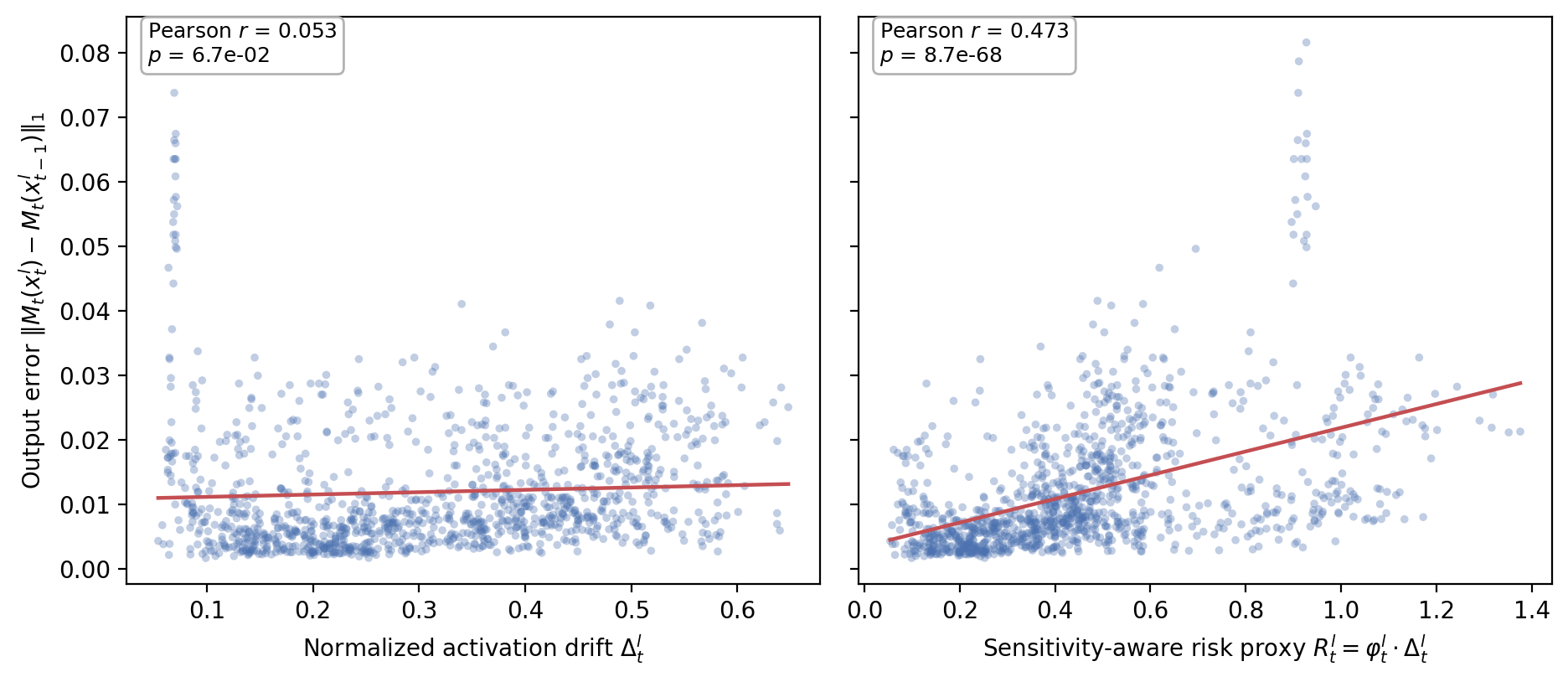}
    \caption{Correlation between the reuse criterion and the true model output error. Left: raw activation drift $\Delta_t^l$; right: sensitivity-aware risk proxy $R_t^l=\varphi_t^l\cdot\Delta_t^l$.}
    \label{fig:scatter_correlation}
\end{figure}

As shown in Figure~\ref{fig:scatter_correlation}, the risk proxy markedly strengthens the correlation with the true output error. Pooled across prompts and layers, the Pearson correlation improves from 0.053 for the raw drift to 0.473 for the risk proxy. Incorporating layer sensitivity thus turns a drift signal that is nearly uncorrelated with the true error into a relatively reliable reuse criterion, which supports the first-order decomposition in Section~3.3.

\subsection[Profiling and Memory Overhead]{Profiling and Memory Overhead}

MG-Cache performs offline calibration once per model; the resulting schedule, importance scores, and regression coefficients are reused across all prompts. On single NVIDIA H100 (80GB) GPU, timestep-drift calibration and linear-regression fitting take about 2 hours (roughly the cost of 50 videos), while the layer-importance profiling takes about 60 hours (roughly the cost of $50$ prompts $\times$ $n/2$ full video generations, where $n$ is the number of DiT layers), since each perturbed forward pass resumes only from the injected layer. The profiling process can be efficiently distributed across multiple GPUs to reduce the profiling time.

The memory overhead is dominated by the layer-wise cache, which stores the reference input and the hidden states of each of the $60$ DiT layers ($2\times60\times L\times3072$ bf16 elements for a sequence of length $L$); the timestep cache ($L\times64$ elements) is negligible. Table~\ref{tab:appendix_memory} reports the cache sizes for the evaluated configurations. When $L$ exceeds $40$k tokens, the layer cache is offloaded to system memory (RAM) with the next layer's cache prefetched to hide transfer latency (Section~4.4).

\begin{table}[tb]
    \centering
    \caption{Memory overhead of MG-Cache caches (bf16) on HunyuanVideo.}
    \label{tab:appendix_memory}
    \footnotesize
    \setlength{\tabcolsep}{2pt}
    \begin{tabular*}{\linewidth}{@{\extracolsep{\fill}}l|c c}
        \hline
         & \textbf{$480\times640$p, 65 frames} & \textbf{$544\times960$p, 129 frames} \\
        \hline
        Sequence length $L$ & 20.7k & 67.6k \\
        Layer cache & 15.2 GB & 49.8 GB \\
        Timestep cache & 2.6 MB & 8.6 MB \\
        Placement & GPU memory & RAM (offloading) \\
        \hline
    \end{tabular*}
\end{table}

\subsection{Sensitivity to $\alpha$ and $\beta$}

\begin{figure}[tb]
    \centering
    \begin{subfigure}{\linewidth}
        \centering
        \includegraphics[width=\linewidth]{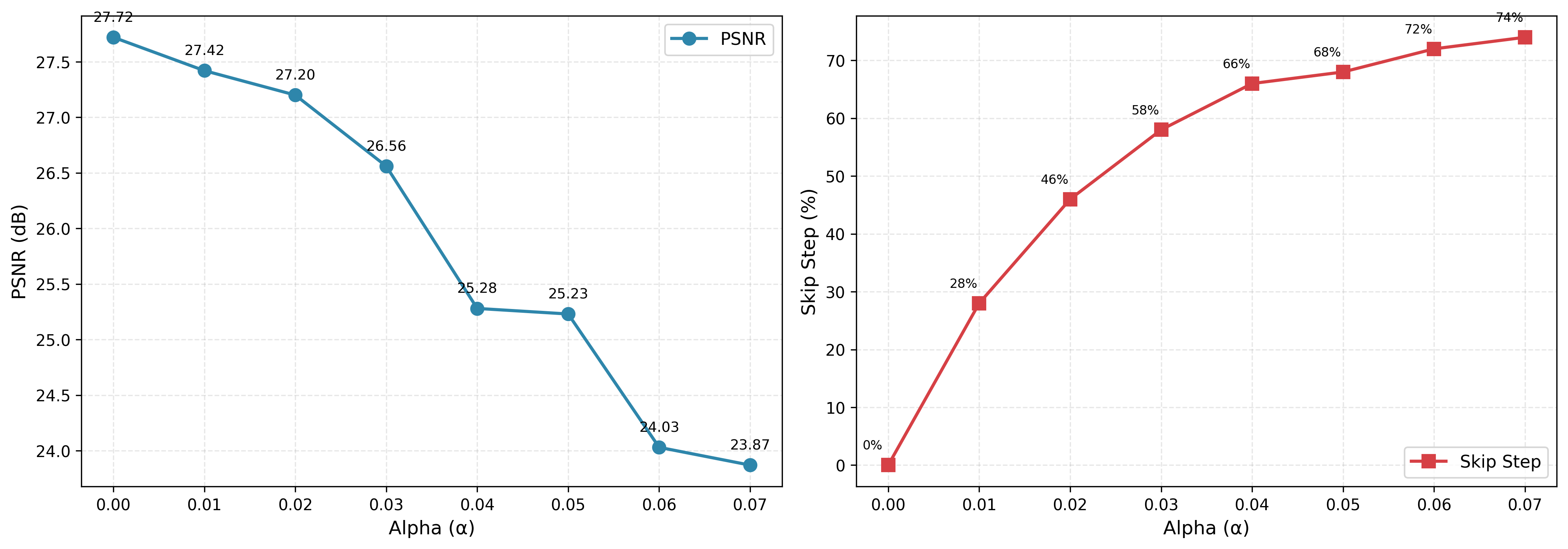}
        \caption{Effect of varying $\alpha$ with fixed $\beta=0.3$.}
    \end{subfigure}

    \begin{subfigure}{\linewidth}
        \centering
        \includegraphics[width=\linewidth]{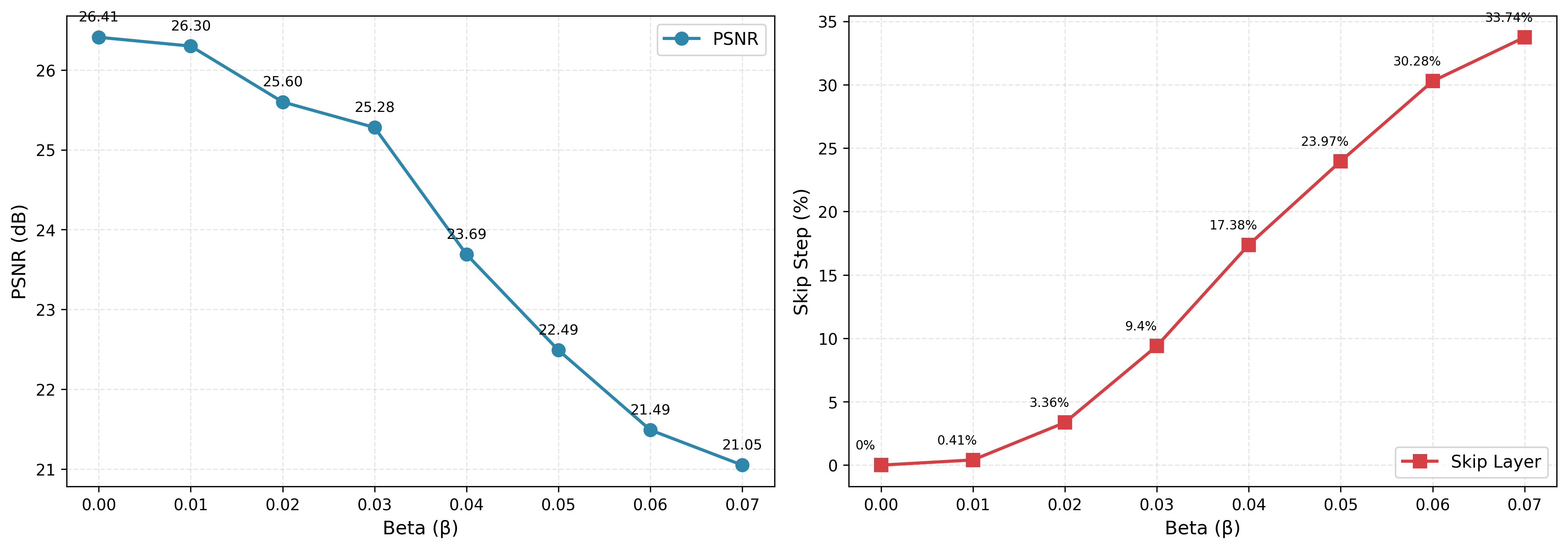}
        \caption{Effect of varying $\beta$ with fixed $\alpha=0.04$.}
    \end{subfigure}
    \caption{\textbf{Additional hyperparameter analysis.} Left/right axes report PSNR and the skip ratio. Larger $\alpha$ ($\beta$) increases timestep (layer) skipping, improving acceleration at the cost of a gradual fidelity drop, with the layer-stage budget exhibiting a steeper trade-off.}
    \label{fig:appendix_alpha_beta}
\end{figure}

We provide additional hyperparameter sweeps for the two error-budget thresholds of MG-Cache. Figure~\ref{fig:appendix_alpha_beta} summarizes how quality and acceleration vary as each hyperparameter changes while the other is fixed.

With $\beta=0.3$ fixed, increasing $\alpha$ raises the timestep skip ratio but gradually lowers PSNR; the trade-off is smooth for small $\alpha$ and steepens as skipping becomes aggressive, supporting the interpretation of $\alpha$ as the coarse-grained timestep risk budget.

With $\alpha=0.04$ fixed, increasing $\beta$ raises the layer skip ratio, and PSNR declines more sharply than in the $\alpha$ sweep, indicating that the layer-stage budget is more sensitive to local approximation error. This is consistent with our main design: timestep skipping removes prompt-robust redundancy first, while layer-wise caching is controlled more carefully through sensitivity-aware risk estimation.

\subsection{Cross-Prompt Stability of Offline Statistics}

To further support the consistency of the offline calibration stage, Table~\ref{tab:appendix_phi_std} reports the standard deviation of layer importance scores $\varphi_t^l$ across different prompts, and Table~\ref{tab:appendix_linear_std} reports the average standard deviation of the linear regression coefficients $K$ and $B$ across prompts. The values remain small across representative timesteps and layers, especially after the earliest denoising stage, which supports the use of shared offline statistics during inference.

The layer-importance results provide additional evidence for the argument in Section~3.3: the prompt dependence of dynamic layer caching mainly comes from the online activation-drift term rather than from the downstream robustness term. Similarly, the low variance of linear regression coefficients shows that the timestep compensation learned offline is stable enough to be reused across prompts without prompt-specific refitting.

\begin{table}[tb]
    \centering
    {\footnotesize
    \setlength{\tabcolsep}{3pt}
    \begin{tabular}{l|c c c c c}
        \hline
        \textbf{Layer / Step} & \textbf{Step-0} & \textbf{Step-10} & \textbf{Step-20} & \textbf{Step-30} & \textbf{Step-40} \\
        \hline
        Layer-5  & 0.0025 & 0.0043 & 0.0050 & 0.0057 & 0.0056 \\
        Layer-15 & 0.0042 & 0.0100 & 0.0095 & 0.0088 & 0.0101 \\
        Layer-25 & 0.0037 & 0.0085 & 0.0079 & 0.0089 & 0.0098 \\
        Layer-35 & 0.0066 & 0.0091 & 0.0081 & 0.0084 & 0.0097 \\
        Layer-45 & 0.0164 & 0.0057 & 0.0061 & 0.0067 & 0.0070 \\
        Layer-55 & 0.0078 & 0.0048 & 0.0048 & 0.0054 & 0.0058 \\
        \hline
    \end{tabular}}
    \caption{Standard deviation of layer importance scores $\varphi_t^l$ across prompts.}
    \label{tab:appendix_phi_std}

    \vspace{\baselineskip}

    \begin{tabular}{l|c c c c c}
        \hline
         & \textbf{Step-0} & \textbf{Step-10} & \textbf{Step-20} & \textbf{Step-30} & \textbf{Step-40} \\
        \hline
        $\mathbf{K}$ & 0.0698 & 0.0013 & 0.0010 & 0.0006 & 0.0006 \\
        $\mathbf{B}$ & 0.1642 & 0.0032 & 0.0017 & 0.0019 & 0.0015 \\
        \hline
    \end{tabular}
    \caption{Average standard deviation of linear regression coefficients across prompts.}
    \label{tab:appendix_linear_std}
\end{table}

\section{Conclusion}

In this study, we propose multi-granularity caching (MG-Cache) designed to accelerate the inference of diffusion transformer models. Our approach contains a two-stage strategy which combines timestep cache and layer-wise cache. The offline timestep analysis is introduced to identify the critical timesteps, and then the remaining timesteps are accelerated with the adaptive layer-wise caching based on layer importance scores and input stability. Finally, we integrate linear regression mechanism to mitigate the errors. Extensive experiments demonstrated that MG-Cache achieves substantial speedups while preserving high-fidelity generation quality. These results validate the effectiveness of MG-Cache in enabling the efficient deployment of high-resolution and long-duration video generation models in practice. This work does not include human perceptual evaluation for assessing visual quality; a systematic human evaluation is left for future work.

\section{Acknowledgments}
We thank all anonymous reviewers for their constructive feedback. This work is supported by the National Natural Science Foundation of China under Grant No. 62372055, the National Science and Technology Major Project (2023ZD0120502), and the Fundamental Research Funds for the Central Universities.

\bibliographystyle{eg-alpha-doi}
\bibliography{references}

\end{document}